\PassOptionsToPackage{unicode}{hyperref}
\PassOptionsToPackage{hyphens}{url}
\documentclass[
]{article}
\usepackage{xcolor}
\usepackage{url}
\usepackage{xcolor}
\usepackage{hyperref}

\hypersetup{
    colorlinks=true,
    urlcolor=blue,
    linkcolor=blue,
    citecolor=blue,
    filecolor=blue
}
\usepackage{hyperref}
\usepackage{amsmath,amssymb}
\usepackage{iftex}
\ifPDFTeX
  \usepackage[T1]{fontenc}
  \usepackage[utf8]{inputenc}
  \usepackage{textcomp} 
\else 
  \usepackage{unicode-math} 
  \defaultfontfeatures{Scale=MatchLowercase}
  \defaultfontfeatures[\rmfamily]{Ligatures=TeX,Scale=1}
\fi
\usepackage{lmodern}
\ifPDFTeX\else
\fi
\IfFileExists{upquote.sty}{\usepackage{upquote}}{}
\IfFileExists{microtype.sty}{
  \usepackage[]{microtype}
  \UseMicrotypeSet[protrusion]{basicmath} 
}{}
\makeatletter
\@ifundefined{KOMAClassName}{
  \IfFileExists{parskip.sty}{%
    \usepackage{parskip}
  }{
    \setlength{\parindent}{0pt}
    \setlength{\parskip}{6pt plus 2pt minus 1pt}}
}{
  \KOMAoptions{parskip=half}}
\makeatother
\usepackage{longtable,booktabs,array}
\usepackage{caption}
\usepackage{calc} 
\usepackage{etoolbox}
\makeatletter
\patchcmd\longtable{\par}{\if@noskipsec\mbox{}\fi\par}{}{}
\makeatother
\IfFileExists{footnotehyper.sty}{\usepackage{footnotehyper}}{\usepackage{footnote}}
\makesavenoteenv{longtable}
\usepackage{graphicx}
\makeatletter
\newsavebox\pandoc@box
\newcommand*\pandocbounded[1]{
  \sbox\pandoc@box{#1}%
  \Gscale@div\@tempa{\textheight}{\dimexpr\ht\pandoc@box+\dp\pandoc@box\relax}%
  \Gscale@div\@tempb{\linewidth}{\wd\pandoc@box}%
  \ifdim\@tempb\p@<\@tempa\p@\let\@tempa\@tempb\fi
  \ifdim\@tempa\p@<\p@\scalebox{\@tempa}{\usebox\pandoc@box}%
  \else\usebox{\pandoc@box}%
  \fi%
}
\def\fps@figure{htbp}
\makeatother
\usepackage{bookmark}
\IfFileExists{xurl.sty}{\usepackage{xurl}}{} 
\makeatletter
\@ifundefined{xmpquote}{}{}
\makeatother
\hypersetup{
  hidelinks,
  pdfcreator={LaTeX via pandoc}}

\author{}
\date{}

\usepackage[margin=1in]{geometry}
\usepackage{lineno}
\usepackage{newunicodechar}
\newunicodechar{γ}{\ensuremath{\gamma}}
\newunicodechar{λ}{\ensuremath{\lambda}}
\newunicodechar{π}{\ensuremath{\pi}}
\newunicodechar{ρ}{\ensuremath{\rho}}
\newunicodechar{τ}{\ensuremath{\tau}}
\newunicodechar{φ}{\ensuremath{\varphi}}
\newunicodechar{≈}{\ensuremath{\approx}}
\newunicodechar{≥}{\ensuremath{\geq}}
\newunicodechar{∈}{\ensuremath{\in}}
\newunicodechar{⟨}{\ensuremath{\langle}}
\newunicodechar{⟩}{\ensuremath{\rangle}}
\newunicodechar{′}{\ensuremath{^{\prime}}}
\newunicodechar{⁻}{\ensuremath{^{-}}}
\newunicodechar{⁴}{\ensuremath{^{4}}}
\newunicodechar{⁷}{\ensuremath{^{7}}}
\newunicodechar{₂}{\ensuremath{_{2}}}

\begin{document}
\begin{center}
\textbf{Matched but Not Better: An Equivalence-Tested Benchmark of
Quantum Machine Learning on Six Clinical Datasets and the
QML-Health-READY Reporting Framework}
\end{center}
\vspace{0.5cm}

\begin{center}
\emph{Mohammed Siraj B\textsuperscript{a}, Gousia Habib \textsuperscript{*} and Ruban S\textsuperscript{a}\\
\textsuperscript{a}School of Engineering, St Aloysius (Deemed to be University), Mangalore
575003, Karnataka, India\\
\textsuperscript{*}School of ICT, Bahrain Polytechnic, Manama, Bahrain.\\
*Corresponding author: Gousia Habib, 
\href{mailto:er.gousiya91@gmail.com}{\textcolor{blue}{er.gousiya91@gmail.com}}}
\end{center}

\section{Abstract}\label{abstract}

\textbf{Background:} Quantum machine learning (QML) is promoted for
clinical prediction, but its comparisons with classical models are
rarely paired or tested for equivalence.\\
\textbf{Methods:} On six public clinical datasets, we compared a
quantum-kernel support vector machine (QSVM; 4 qubits, tuned bandwidth)
and a variational quantum classifier (VQC) with five tuned classical
models, using identical four-component inputs and 25 identical
cross-validation splits. Paired differences in the area under the ROC
curve (AUROC) were assessed with a corrected resampled t-test,
Holm-adjusted equivalence tests (margins $\pm$0.02 and $\pm$0.05) and
random-effects pooling. Ablation, noise, finite-shot and
implementation-fault analyses followed. We propose QML-Health-READY, an
eight-domain reporting framework.\\
\textbf{Results:} No quantum model significantly outperformed any
classical comparator. Pooled across datasets, QSVM differed from the
kernel-matched radial basis function (RBF) SVM by -0.006 AUROC (95\% CI
-0.017 to +0.004; equivalent within $\pm0.02$), and the VQC differed from
the best classical model by -0.007 (-0.019 to +0.004). Per dataset, QSVM
was equivalent to the RBF SVM within $\pm0.05$ on 4 of 6 but significantly
worse than the best classical model on NHANES 2017--18 (-0.076). A
classically trivial product-state kernel performed as well as the
entangling kernel. With p = 0.01 depolarising noise and 1,024 shots per
kernel entry, QSVM AUROC fell by a pooled 0.054 (0.036--0.071); an
untuned bandwidth gave near-chance AUROC (0.48--0.59).\\
\textbf{Conclusions:} Correctly configured quantum classifiers
approached, but did not exceed, tuned classical models; finite-shot
estimation, not gate noise, was their main practical cost, and
configuration errors alone could reverse the conclusion.
QML-Health-READY specifies the reporting that credible claims require.

\textbf{Keywords:} \textbf{Quantum machine learning; Quantum kernel; Equivalence
testing; Benchmarking, Reporting guideline; Clinical prediction}

\section{1. Introduction}\label{introduction}

Machine learning is now embedded in clinical research and practice
{[}1,2{]}. Noisy intermediate-scale quantum (NISQ) hardware {[}3{]} has
prompted parallel interest in quantum machine learning (QML), in which a
parameterised quantum circuit serves as a feature map or kernel
{[}4--6{]}. The quantum support vector machine (QSVM) was proposed with
a logarithmic-runtime claim {[}7{]}, and the exponentially large Hilbert
space has been argued to offer expressive power unavailable to classical
models {[}8,9{]}. For clinical prediction the question is practical:
whether a quantum model discriminates better than a well-tuned classical
one, at a cost a health system could bear. Overstated claims can divert
research effort and funding from methods that already work.

The evidence has not followed. A 2025 systematic review identified 4,915
records (2015 to June 2024), evaluated 46 studies of QML for digital
health in detail after excluding many for insufficient rigour, found
that only 16 tested algorithms under realistic conditions (quantum
hardware or noisy circuits), and concluded that there is not yet clear
evidence of an empirical advantage {[}10{]}. A benchmark of 12 QML
models on 160 datasets found that out-of-the-box classical models
outperformed the quantum classifiers {[}11{]}, earlier reviews were
similarly cautious {[}12{]}, and dequantisation results show that
classical algorithms can often match proposed quantum speed-ups or
trained quantum models {[}13,14{]}. Independent benchmarks from 2025 and
2026 point the same way: quantum-kernel SVMs on nine tabular datasets,
validated on IBM hardware, showed no significant advantage in any of 29
comparisons {[}15{]}, and a multi-modal oncology benchmark found no
evidence of quantum advantage {[}16{]}. Applied medical studies
nevertheless continue to appear, on hardware {[}17{]} and in simulation
{[}18{]}, and recent reviews document highly heterogeneous methods
{[}19{]}.

Several mechanisms behind these results are now understood. Quantum
fidelity kernels concentrate exponentially with qubit count {[}20{]};
performance depends strongly on a bandwidth hyperparameter {[}21,22{]},
and once that bandwidth is tuned the kernels become close to classical
RBF or low-order polynomial kernels {[}23{]}; and concentration and
spread effects inflate the number of measurement shots that kernel
estimation requires {[}24{]}. These results come mostly from physics and
machine-learning benchmarks. What is missing for clinical prediction is
evidence of three kinds: whether quantum and classical models are
equivalent, rather than merely ``not significantly different'', under
identical inputs and splits; how large the shot cost is on clinical data
once bandwidth is tuned; and how far implementation choices alone can
move the conclusion. Our contributions are: (i) a synthesis of seven
reasons why medical QML advantage claims are fragile; (ii)
QML-Health-READY, an eight-domain reporting and readiness framework
aligned with TRIPOD+AI, CLAIM and FUTURE-AI {[}25--27{]}; (iii) a
matched, paired, equivalence-tested benchmark on six clinical datasets
with a kernel-matched classical comparator; (iv) an empirical
quantification of the finite-shot penalty and of its interaction with
bandwidth tuning; and (v) an implementation-sensitivity experiment
showing how plausible configuration faults reverse the conclusion.

\section{2. Why quantum advantage claims in medical QML are
fragile}\label{why-advantage-claims-in-medical-qml-are-fragile}
Claims of quantum advantage in medical machine learning hinge on a chain of assumptions concerning the model class, the structure of the data, the fairness of the comparison, and the cost of executing the model on real hardware. That chain is stressed by medical datasets that are often tiny, noisy, unbalanced and high-dimensional. We thus explore seven sources of fragility, from theoretical bases to experimental practice and hardware limits, and note for each what a plausible claim would need to verify.\\
\textbf{Quantum models are kernel methods (domain A).} Supervised
QML models that embed data in Hilbert space and measure a linear
observable are kernel methods {[}5,28{]}. Whether a quantum kernel beats
an RBF or polynomial kernel is an empirical, data-dependent question,
and expressive power depends on the encoding {[}29,30{]}.\\
\textbf{Advantage depends on data structure (domain R).} A quantum
learning advantage requires a large geometric difference between quantum
and classical kernels on the data at hand {[}31{]}; without such an
analysis an advantage claim has no theoretical basis.\\
\textbf{Dequantisation (domain Y).} Quantum-inspired classical
algorithms have matched several proposed speed-ups {[}13{]},
random-Fourier-feature models can approximate trained quantum kernel and
variational models under stated conditions {[}14{]}, and the regimes in
which quantum and classical models make different predictions are being
characterised {[}32{]}.\\
\textbf{Benchmarking practice (domain A).} Results are highly
sensitive to experimental design {[}11,33{]}. A tuned quantum model
compared with an untuned classical model, or one given different inputs,
yields uninformative results.\\
\textbf{Trainability and noise (domain R).} Variational circuits can
exhibit barren plateaus {[}34,35{]}, noise induces similar effects
{[}36{]}, and many models that provably avoid barren plateaus can be
simulated classically once data are collected from the device {[}37{]}.\\
\textbf{Data loading and clinical realism (domain D).} Amplitude
encoding of an n-dimensional vector into log₂ n qubits needs on the
order of n gates --- exponential in the number of qubits --- which
removes logarithmic speed-ups {[}38{]}. Angle encoding is tractable but
forces strong dimensionality reduction, so classical comparators must
receive the same reduced inputs or the asymmetry must be reported.\\
\textbf{Kernel scaling, bandwidth and concentration (domain E).}
Fidelity kernels concentrate exponentially with qubit count, making Gram
matrices nearly diagonal {[}20{]}. Performance depends strongly on a
bandwidth hyperparameter {[}21,22{]}; at the small bandwidths that
generalise well, quantum kernels become close to RBF or low-order
polynomial kernels {[}23,39{]}, and estimating nearly constant kernel
entries demands many measurement shots {[}24{]}. Untuned encodings can
therefore produce spurious failures, and tuned encodings can produce
results that are strong but classically approximable and costly to
measure.\\

\section{3. The QML-Health-READY
framework}\label{the-qml-health-ready-framework}

QML-Health-READY has eight domains (R, E, A, D, Y; R, E, Y), each mapped
to the failure modes of Section 2 (Table 1). Items are answered Yes,
Partial, No or Not applicable with a one-line justification. Studies
meeting all Core items (†) are READY-Reportable; readiness tiers are
defined in Supplementary Table S1, and our own self-assessment is in
Supplementary Table S6. Meeting the Core items is necessary for credible
reporting and does not imply clinical usefulness. Section 5.9 reports,
for each domain, an effect measured in this study that the corresponding
items would have detected.

\emph{Table 1. QML-Health-READY domains. † Core item.}

{\def\LTcaptype{none} 
\begin{longtable}[]{@{}
  >{\raggedright\arraybackslash}p{(\linewidth - 4\tabcolsep) * \real{0.1000}}
  >{\raggedright\arraybackslash}p{(\linewidth - 4\tabcolsep) * \real{0.2167}}
  >{\raggedright\arraybackslash}p{(\linewidth - 4\tabcolsep) * \real{0.6833}}@{}}
\toprule\noalign{}
\begin{minipage}[b]{\linewidth}\raggedright
\textbf{Domain}
\end{minipage} & \begin{minipage}[b]{\linewidth}\raggedright
\textbf{Name}
\end{minipage} & \begin{minipage}[b]{\linewidth}\raggedright
\textbf{Items}
\end{minipage} \\
\midrule\noalign{}
\endhead
\bottomrule\noalign{}
\endlastfoot
R† & Rationale & Hypothesised mechanism for advantage; data-complexity
analysis (geometric difference, kernel alignment) {[}31,40{]}. \\
E† & Encoding & Encoding, feature scaling, kernel bandwidth and its
selection procedure (checking grid edges), dimensionality reduction
fitted within training folds, qubit count {[}21,23{]}. \\
A† & Algorithm \& baselines & ≥ 4 tuned classical baselines including a
kernel-matched classical model, given the same inputs and tuning budget;
entanglement-ablated quantum model {[}11{]}. \\
D† & Data \& splits & Provenance, sample size, class balance,
subject-level splitting, leakage audit (including post-baseline
predictors). \\
Y† & Yield & Repeated nested cross-validation with identical splits for
all models; paired tests corrected for split dependence {[}41,42{]};
equivalence testing with a pre-specified margin when parity is claimed
{[}43,44{]}; multiplicity control. \\
R† & Robustness & Explicit noise model (verified to be active) and
finite-shot estimation; hardware-calibrated noise or real hardware where
feasible {[}20,36{]}. \\
E & Evaluation in context & Calibration, threshold-dependent metrics,
decision-curve analysis, external validation; alignment with TRIPOD+AI,
CLAIM and FUTURE-AI {[}25--27{]}. \\
Y & Yardstick for cost & Wall-clock time, circuit executions and shots
per fit versus classical baselines; data-access assumptions;
dequantisation check {[}13,14{]}. \\
\end{longtable}
}

\section{4. Methods}\label{methods}
The study is a matched, paired benchmark of two quantum classifiers
against five tuned classical models on six public clinical datasets,
with all quantum results obtained by classical simulation. Sections
\ref{sec:datasets}--\ref{sec:complexity} describe the data, models,
statistical analysis and kernel diagnostics.
\subsection{4.1 Datasets}\label{datasets}
We used six public datasets (Table 2): Breast Cancer Wisconsin (benign
coded as class 1, as distributed; AUROC is invariant to this choice);
Heart Disease Cleveland (processed UCI file, incomplete rows removed,
any disease positive); Pima Indians Diabetes (zero values for glucose,
blood pressure, skin thickness, insulin and BMI treated as missing and
median-imputed within each training fold); Heart Failure Clinical
Records {[}45{]} with follow-up time removed because it is recorded
after baseline; TCGA-BRCA oestrogen-receptor status {[}46,47{]} from
Illumina HiSeq expression percentiles, restricted to primary tumours
with one sample per patient (all 780 ER-labelled samples met this
criterion) and to the 200 highest-variance genes, a label-free filter;
and NHANES 2017--18 adults {[}48{]}, with diabetes defined as HbA1c $\geq$
6.5\% or a physician diagnosis and HbA1c excluded from the predictors to
avoid target leakage (a stratified random sample of 2,000 of 4,483
eligible adults with complete data; survey weights not applied).

\emph{Table 2. Datasets. Prevalence is the fraction of class 1, the
positive class for AUROC.}

{\def\LTcaptype{none} 
\begin{longtable}[]{@{}
  >{\raggedright\arraybackslash}p{(\linewidth - 12\tabcolsep) * \real{0.1428}}
  >{\raggedright\arraybackslash}p{(\linewidth - 12\tabcolsep) * \real{0.1431}}
  >{\raggedright\arraybackslash}p{(\linewidth - 12\tabcolsep) * \real{0.1428}}
  >{\raggedright\arraybackslash}p{(\linewidth - 12\tabcolsep) * \real{0.1428}}
  >{\raggedright\arraybackslash}p{(\linewidth - 12\tabcolsep) * \real{0.1428}}
  >{\raggedright\arraybackslash}p{(\linewidth - 12\tabcolsep) * \real{0.1428}}
  >{\raggedright\arraybackslash}p{(\linewidth - 12\tabcolsep) * \real{0.1428}}@{}}
\toprule\noalign{}
\begin{minipage}[b]{\linewidth}\raggedright
\textbf{Dataset}
\end{minipage} & \begin{minipage}[b]{\linewidth}\raggedright
\textbf{Domain}
\end{minipage} & \begin{minipage}[b]{\linewidth}\raggedright
\textbf{Source}
\end{minipage} & \begin{minipage}[b]{\linewidth}\raggedright
\textbf{n}
\end{minipage} & \begin{minipage}[b]{\linewidth}\raggedright
\textbf{Predictors}
\end{minipage} & \begin{minipage}[b]{\linewidth}\raggedright
\textbf{Class 1}
\end{minipage} & \begin{minipage}[b]{\linewidth}\raggedright
\textbf{Prevalence}
\end{minipage} \\
\midrule\noalign{}
\endhead
\bottomrule\noalign{}
\endlastfoot
Breast Cancer Wisconsin & Oncology & scikit-learn / UCI & 569 & 30 &
benign tumour & 63\% \\
Heart Disease Cleveland & Cardiology & UCI & 297 & 13 & any heart
disease & 46\% \\
Pima Indians Diabetes & Metabolic & UCI (mirror) & 768 & 8 & diabetes &
35\% \\
Heart Failure & Cardiology & UCI & 299 & 11 & death during follow-up &
32\% \\
TCGA-BRCA & Genomic oncology & UCSC Xena & 780 & 200 & ER-positive &
77\% \\
NHANES 2017--18 & Epidemiology & CDC NHANES & 2,000 & 6 & diabetes &
18\% \\
\end{longtable}
}

\subsection{4.2 Models}\label{models}

Within each outer training fold we fitted median imputation,
standardisation, principal component analysis (PCA) to 4 components and
re-standardisation of the components; all models received exactly these
four inputs, and the classical models were additionally fitted on all
standardised predictors to quantify the cost of the reduction. Classical
models (L2 logistic regression, support vector machines (SVMs) with
radial basis function (RBF) and polynomial kernels, random forest,
gradient boosting) were tuned by 3-fold inner grid search on AUROC
{[}49{]} (grids in Supplementary Table S5).

The QSVM uses the ZZ feature map {[}4{]} with 2 repetitions on a linear
chain: Hadamard and $R_z(2bz_i)$ on each qubit, then
CNOT--$R_z\bigl(2(\pi-bz_i)(\pi-bz_{i+1})\bigr)$--CNOT, where $z$ are the standardised
components and $b$ is a bandwidth {[}21,22{]}. The fidelity kernel
$|\langle\phi(x')|\phi(x)\rangle|^2$ was computed by exact state-vector
simulation in PennyLane 0.45.1 {[}50{]}. Bandwidth (0.005, 0.01, 0.02,
0.05, 0.1, 0.2, 0.5, 1.0) and C (0.1, 1, 10) were tuned by 3-fold inner
cross-validation on the AUROC of the SVM decision function, and the SVM
was refitted on the full training fold. The product-state ablation
removes the CNOT blocks. The noisy QSVM applies a depolarising channel
(p = 0.01) to every qubit acted on by every gate (density-matrix
simulation) and uses $\mathrm{Tr}\bigl(\rho(x)\rho(x')\bigr)$ as the kernel, re-tuning $b$ and $C$; a
finite-shot arm additionally estimates every kernel entry from 1,024
shots and projects the training Gram matrix onto the positive
semidefinite cone. Shots were modelled as binomial draws with success
probability equal to the exact kernel value; this is exact for the
noiseless compute--uncompute estimator but only approximate for the
noisy kernel, and two-qubit gates received independent single-qubit
channels rather than a two-qubit channel. Because global depolarising
noise maps the kernel to a·K + c, which an SVM with a bias term and
re-tuned C largely absorbs, exact noisy kernels are expected to change
AUROC little; finite-shot estimation is the more demanding test
{[}20{]}. A noise sweep (p = 0--0.1) used the first seed's folds at the
bandwidth and C selected for the noiseless QSVM, as did a shot sweep.

The VQC applies AngleEmbedding (Rx) and 3 StronglyEntanglingLayers
{[}34,51,52{]}, reads out ⟨Z⟩ on all qubits through a trainable linear
layer, and was trained by full-batch Adam (learning rate 0.05, 100
epochs, cross-entropy); its number of layers and learning rate were
tuned by 3-fold inner cross-validation over \{(1, 0.05), (3, 0.05), (5,
0.05), (3, 0.15)\}, as for the QSVM, and the fixed configuration (3
layers, learning rate 0.05) is reported as a sensitivity analysis. Its
ablation replaces the entangling layers with single-qubit rotations;
with all-qubit readout this removes entanglement only.

To quantify implementation sensitivity we re-ran the QSVM with three
cumulative configuration faults, all present in an earlier version of
our own pipeline and plausible elsewhere: bandwidth fixed at 1; plus
unscaled principal components as circuit inputs; plus a CZ--Rz--CZ
entangling block, which equals a single-qubit Rz because diagonal gates
commute, so the circuit creates no entanglement. We also ran a widely
used tutorial-style configuration, with inputs min--max scaled to {[}0,
π{]} on each training fold and no bandwidth. A qubit-scaling experiment
(4, 8 and 12 qubits; first-seed folds) on TCGA-BRCA and breast cancer
compared the QSVM with an RBF SVM on the same components and tracked
kernel concentration at bandwidth 1.

\subsection{4.3 Statistical analysis}\label{statistical-analysis}

Five seeds × 5-fold stratified cross-validation gave 25 outer splits per
dataset, identical for every model, so all comparisons are paired. The
primary metric is the area under the receiver operating characteristic
curve (AUROC). Because overlapping training sets make the 25 differences
dependent, we used the corrected resampled t-test (variance inflated by
1/J + n\_test/n\_train) {[}41,42{]} and, for parity claims, two
one-sided tests (TOST) {[}43,44{]} with the same variance at margins of
$\pm$0.01, $\pm$0.02 and $\pm$0.05 AUROC. Within each comparison family, p-values
across the six datasets were Holm-adjusted {[}53{]}. A comparison is
Equivalent at a margin if its adjusted TOST p \textless{} 0.05, Worse or
Better if its adjusted t-test p \textless{} 0.05, and otherwise
Inconclusive. To summarise across datasets, the per-dataset differences
and their corrected standard errors were pooled with a random-effects
model (DerSimonian--Laird heterogeneity; Hartung--Knapp--Sidik--Jonkman
confidence interval with $k-1$ degrees of freedom) {[}54,55{]}, and
pooled equivalence was declared when the 90\% confidence interval lay
inside the margin. With six datasets these pooled estimates summarise
the datasets studied rather than all clinical prediction tasks. The
primary QSVM comparison is with the RBF SVM, which differs only in the
kernel and, like the QSVM, was scored by its decision function;
comparisons with the best classical model per dataset (selected post
hoc, which favours the classical side) and for the VQC are secondary.
Accuracy, macro-F1, balanced accuracy and the Matthews correlation
coefficient (MCC) are reported in Supplementary Table S4; thresholds
differ between models (probability 0.5 versus decision-function sign),
so these are descriptive only.

The analysis was not pre-registered, and parts of it were revised after
an initial analysis of an earlier version of the pipeline (Supplementary
Table S8). The $\pm$0.05 margin was specified before any results were seen.
The $\pm$0.01 and $\pm$0.02 margins, the designation of the RBF SVM as the
primary comparator, and the finite-shot, all-predictor,
implementation-sensitivity and qubit-scaling analyses were added
afterwards. All margins are conventional rather than clinically derived,
and all comparisons that were run are reported. Because this is a
methodological benchmark rather than the development of a clinical
prediction tool, we do not claim TRIPOD+AI compliance; the items it
covers that apply here (data sources, outcome and predictor definitions,
missing data, validation design and performance measures) are reported
in Sections 4.1--4.3. A complete independent re-run of the benchmark
reproduced every per-split AUROC exactly except those of SVMs scored
with Platt-scaled probabilities, which rely on an internal random split
and differed by at most 0.001 per split; this is why the kernel-matched
comparison scores the RBF SVM by its decision function.

\subsection{4.4 Kernel and data-complexity
analysis}\label{kernel-and-data-complexity-analysis}

On a stratified sample of up to 300 observations we computed the maximum
per-feature Fisher ratio, 1-nearest-neighbour label agreement, centred
kernel--target alignment {[}40{]} for the quantum, product-state and RBF
kernels, and the geometric difference g(K\_C‖K\_Q) {[}31{]} from
trace-normalised kernels with regularisation λ = 10-³ (and 10-⁴ to 10-¹
as a sensitivity analysis), together with the centred alignment between
the quantum kernel and the best-matching RBF kernel. $\sqrt{n}$ is used as a
reference scale: Huang et al. relate a potential advantage to g
approaching $\sqrt{n}$, but g is not bounded by $\sqrt{n}$ under regularisation.
Provable speed-ups in supervised learning require data-access
assumptions that clinical datasets do not satisfy {[}38{]}.

\section{5. Results}\label{results}
We first report discrimination for all models on identical four-component
inputs (5.1), then the paired and pooled
comparisons with equivalence tests (5.2), which form
the primary analysis: the QSVM against the kernel-matched RBF SVM, with
comparisons against the best classical model and the VQC as secondary.
We then examine whether entanglement matters (5.3), how noise and finite
shots affect the QSVM (5.4), and the kernel regime and data complexity
that explain the results (5.5). The remaining subsections report
implementation sensitivity (5.6), the effect of dimensionality reduction
(5.7), computational cost (5.8) and the measured effect behind each
QML-Health-READY domain (5.9).
\subsection{5.1 Discrimination}\label{discrimination}

Discrimination varied widely across datasets (Table 3; Figure 1). The
best classical model reached AUROC 0.736 (Heart Failure) to 0.994
(Breast Cancer Wisconsin), and no single classical model was best
everywhere (SVM (RBF) on 3; logistic regression on 2; SVM (polynomial)
on 1). The noiseless QSVM ranged from 0.700 to 0.990 and tracked the
classical models closely on every dataset except NHANES 2017--18. Among
four-component models, a quantum variant had the highest mean AUROC on
Pima Indians Diabetes and NHANES 2017--18 (QSVM (product-state kernel),
VQC), but by at most 0.001 AUROC over the best classical model, far
inside the confidence intervals.

\emph{Table 3. AUROC, mean $\pm$ 95\% CI over 25 paired splits (CIs treat
splits as independent and are therefore optimistic). Bold: highest mean
per dataset among four-component models. The last row gives the best
classical model on all predictors.}

{\def\LTcaptype{none} 
\begin{longtable}[]{@{}
  >{\raggedright\arraybackslash}p{(\linewidth - 12\tabcolsep) * \real{0.2000}}
  >{\raggedright\arraybackslash}p{(\linewidth - 12\tabcolsep) * \real{0.1333}}
  >{\raggedright\arraybackslash}p{(\linewidth - 12\tabcolsep) * \real{0.1333}}
  >{\raggedright\arraybackslash}p{(\linewidth - 12\tabcolsep) * \real{0.1333}}
  >{\raggedright\arraybackslash}p{(\linewidth - 12\tabcolsep) * \real{0.1333}}
  >{\raggedright\arraybackslash}p{(\linewidth - 12\tabcolsep) * \real{0.1333}}
  >{\raggedright\arraybackslash}p{(\linewidth - 12\tabcolsep) * \real{0.1333}}@{}}
\toprule\noalign{}
\begin{minipage}[b]{\linewidth}\raggedright
\textbf{Model}
\end{minipage} & \begin{minipage}[b]{\linewidth}\raggedright
\textbf{Breast Cancer Wisconsin}
\end{minipage} & \begin{minipage}[b]{\linewidth}\raggedright
\textbf{Heart Disease Cleveland}
\end{minipage} & \begin{minipage}[b]{\linewidth}\raggedright
\textbf{Pima Indians Diabetes}
\end{minipage} & \begin{minipage}[b]{\linewidth}\raggedright
\textbf{Heart Failure}
\end{minipage} & \begin{minipage}[b]{\linewidth}\raggedright
\textbf{TCGA-BRCA}
\end{minipage} & \begin{minipage}[b]{\linewidth}\raggedright
\textbf{NHANES 2017--18}
\end{minipage} \\
\midrule\noalign{}
\endhead
\bottomrule\noalign{}
\endlastfoot
Logistic regression & 0.993 $\pm$ 0.002 & \textbf{0.898 $\pm$ 0.022} & 0.815 $\pm$
0.013 & 0.710 $\pm$ 0.029 & 0.954 $\pm$ 0.009 & 0.777 $\pm$ 0.012 \\
SVM (RBF) & \textbf{0.994 $\pm$ 0.002} & 0.892 $\pm$ 0.022 & 0.820 $\pm$ 0.013 &
\textbf{0.736 $\pm$ 0.028} & 0.952 $\pm$ 0.010 & 0.731 $\pm$ 0.015 \\
SVM (polynomial) & 0.993 $\pm$ 0.004 & 0.889 $\pm$ 0.022 & 0.817 $\pm$ 0.014 & 0.712
$\pm$ 0.019 & \textbf{0.958 $\pm$ 0.008} & 0.712 $\pm$ 0.020 \\
Random forest & 0.985 $\pm$ 0.004 & 0.880 $\pm$ 0.023 & 0.812 $\pm$ 0.014 & 0.709 $\pm$
0.027 & 0.946 $\pm$ 0.011 & 0.773 $\pm$ 0.012 \\
Gradient boosting & 0.985 $\pm$ 0.004 & 0.870 $\pm$ 0.022 & 0.812 $\pm$ 0.014 &
0.664 $\pm$ 0.025 & 0.944 $\pm$ 0.008 & 0.770 $\pm$ 0.012 \\
QSVM & 0.990 $\pm$ 0.003 & 0.884 $\pm$ 0.024 & 0.807 $\pm$ 0.014 & 0.708 $\pm$ 0.028 &
0.956 $\pm$ 0.010 & 0.700 $\pm$ 0.016 \\
QSVM (noisy, p = 0.01) & 0.987 $\pm$ 0.003 & 0.878 $\pm$ 0.024 & 0.804 $\pm$ 0.014 &
0.710 $\pm$ 0.032 & 0.956 $\pm$ 0.010 & 0.704 $\pm$ 0.017 \\
QSVM (noisy + 1,024 shots) & 0.942 $\pm$ 0.008 & 0.808 $\pm$ 0.040 & 0.738 $\pm$
0.021 & 0.630 $\pm$ 0.031 & 0.918 $\pm$ 0.015 & 0.611 $\pm$ 0.011 \\
QSVM (product-state kernel) & 0.993 $\pm$ 0.002 & 0.897 $\pm$ 0.023 &
\textbf{0.821 $\pm$ 0.013} & 0.735 $\pm$ 0.026 & 0.953 $\pm$ 0.010 & 0.728 $\pm$
0.018 \\
VQC & 0.991 $\pm$ 0.002 & 0.873 $\pm$ 0.022 & 0.809 $\pm$ 0.013 & 0.686 $\pm$ 0.027 &
0.954 $\pm$ 0.010 & \textbf{0.778 $\pm$ 0.013} \\
VQC (no entanglement) & 0.988 $\pm$ 0.004 & 0.883 $\pm$ 0.024 & 0.810 $\pm$ 0.013 &
0.727 $\pm$ 0.027 & 0.953 $\pm$ 0.010 & 0.772 $\pm$ 0.013 \\
VQC (tuned) & 0.991 $\pm$ 0.002 & 0.872 $\pm$ 0.022 & 0.807 $\pm$ 0.013 & 0.682 $\pm$
0.026 & 0.955 $\pm$ 0.010 & 0.777 $\pm$ 0.012 \\
SVM (RBF), decision-function score & 0.994 $\pm$ 0.002 & 0.892 $\pm$ 0.022 &
0.820 $\pm$ 0.013 & 0.736 $\pm$ 0.028 & 0.952 $\pm$ 0.009 & 0.731 $\pm$ 0.015 \\
Best classical, all predictors & 0.996 (SVM (polynomial)) & 0.901
(Random forest) & 0.845 (SVM (polynomial)) & 0.781 (Random forest) &
0.961 (Random forest) & 0.787 (Random forest) \\
\end{longtable}

\includegraphics[width=0.9\linewidth,keepaspectratio]{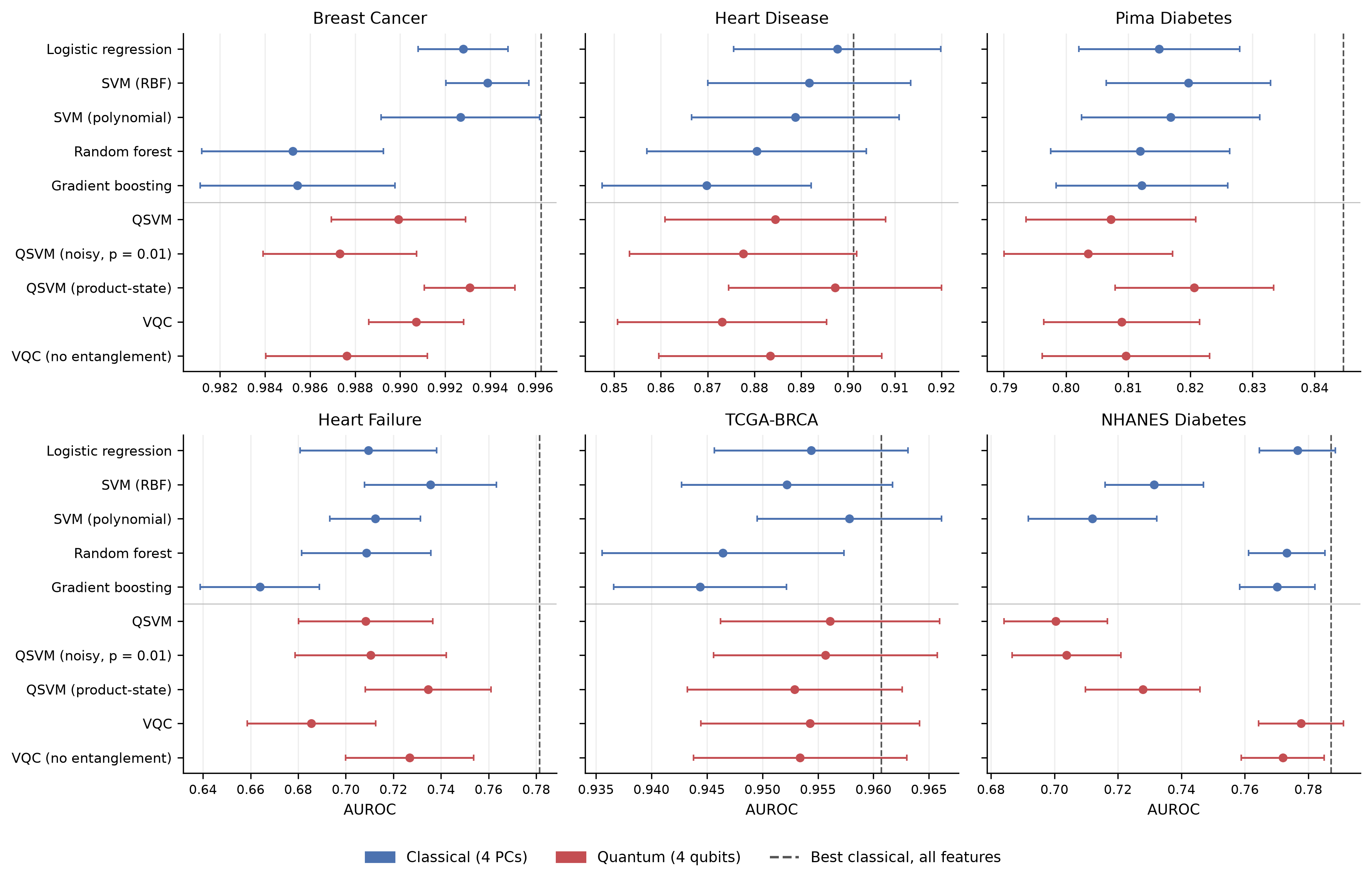}

\emph{Figure 1. AUROC (mean $\pm$ 95\% CI, 25 splits) for classical (blue)
and quantum (red) models on identical four-component inputs. Dashed
line: best classical model on all predictors.}

\subsection{5.2 Paired comparisons and
equivalence}\label{paired-comparisons-and-equivalence}

\emph{Table 4. Paired AUROC differences (quantum-classical) with
corrected 95\% CIs, Holm-adjusted p-values for the corrected t-test and
for equivalence (TOST) at $\pm$0.02 and $\pm$0.05, and verdicts at each margin
(E = equivalent,I = inconclusive,W = worse, B = better).}

\small
{\def\LTcaptype{none} 
\begin{longtable}[]{@{}
  >{\raggedright\arraybackslash}p{(\linewidth - 12\tabcolsep) * \real{0.1583}}
  >{\raggedright\arraybackslash}p{(\linewidth - 12\tabcolsep) * \real{0.2417}}
  >{\raggedright\arraybackslash}p{(\linewidth - 12\tabcolsep) * \real{0.2167}}
  >{\raggedright\arraybackslash}p{(\linewidth - 12\tabcolsep) * \real{0.0833}}
  >{\raggedright\arraybackslash}p{(\linewidth - 12\tabcolsep) * \real{0.1000}}
  >{\raggedright\arraybackslash}p{(\linewidth - 12\tabcolsep) * \real{0.1000}}
  >{\raggedright\arraybackslash}p{(\linewidth - 12\tabcolsep) * \real{0.1000}}@{}}
\toprule\noalign{}
\begin{minipage}[b]{\linewidth}\raggedright
\textbf{Dataset}
\end{minipage} & \begin{minipage}[b]{\linewidth}\raggedright
\textbf{Comparison}
\end{minipage} & \begin{minipage}[b]{\linewidth}\raggedright
\textbf{$\Delta$ AUROC {[}95\% CI{]}}
\end{minipage} & \begin{minipage}[b]{\linewidth}\raggedright
\textbf{p(t)}
\end{minipage} & \begin{minipage}[b]{\linewidth}\raggedright
\textbf{p(TOST $\pm$0.02)}
\end{minipage} & \begin{minipage}[b]{\linewidth}\raggedright
\textbf{p(TOST $\pm$0.05)}
\end{minipage} & \begin{minipage}[b]{\linewidth}\raggedright
\textbf{Verdict $\pm$0.02 / $\pm$0.05}
\end{minipage} \\
\midrule\noalign{}
\endhead
\bottomrule\noalign{}
\endlastfoot
Breast Cancer Wisconsin & QSVM-SVM (RBF) & -0.004 {[}-0.011, +0.003{]}
& 0.923 & \textless0.001 & \textless0.001 & E / E \\
Breast Cancer Wisconsin & QSVM-best (SVM (RBF)) &-0.004 {[}-0.011,
+0.003{]} & 0.569 & \textless0.001 & \textless0.001 & E / E \\
Breast Cancer Wisconsin & VQC (tuned)-best (SVM (RBF)) & -0.003
{[}-0.006, +0.001{]} & 0.473 & \textless0.001 & \textless0.001 & E /
E \\
Heart Disease Cleveland & QSVM-SVM (RBF) & -0.007 {[}-0.034, +0.019{]}
& 1.000 & 0.658 & 0.004 & I / E \\
Heart Disease Cleveland & QSVM - best (Logistic regression) & -0.013
{[}-0.034, +0.007{]} & 0.569 & 0.750 & 0.002 & I / E \\
Heart Disease Cleveland & VQC (tuned) - best (Logistic regression) &
-0.026 {[}-0.047, -0.004{]} & 0.136 & 1.000 & 0.029 & I / E \\
Pima Indians Diabetes & QSVM - SVM (RBF) & -0.012 {[}-0.029, +0.004{]} &
0.626 & 0.658 & \textless0.001 & I / E \\
Pima Indians Diabetes & QSVM - best (SVM (RBF)) & -0.012 {[}-0.029,
+0.004{]} & 0.499 & 0.693 & \textless0.001 & I / E \\
Pima Indians Diabetes & VQC (tuned) - best (SVM (RBF)) & -0.013
{[}-0.029, +0.003{]} & 0.473 & 0.526 & \textless0.001 & I / E \\
Heart Failure & QSVM - SVM (RBF) & -0.027 {[}-0.054, -0.001{]} & 0.259 &
1.000 & 0.087 & I / I \\
Heart Failure & QSVM - best (SVM (RBF)) & -0.027 {[}-0.054, -0.001{]} &
0.213 & 1.000 & 0.087 & I / I \\
Heart Failure & VQC (tuned) - best (SVM (RBF)) & -0.053 {[}-0.110,
+0.003{]} & 0.323 & 1.000 & 0.546 & I / I \\
TCGA-BRCA & QSVM - SVM (RBF) & +0.004 {[}-0.008, +0.016{]} & 1.000 &
0.031 & \textless0.001 & E / E \\
TCGA-BRCA & QSVM - best (SVM (polynomial)) & -0.002 {[}-0.014, +0.011{]}
& 0.777 & 0.016 & \textless0.001 & E / E \\
TCGA-BRCA & VQC (tuned) - best (SVM (polynomial)) & -0.003 {[}-0.016,
+0.010{]} & 1.000 & 0.027 & \textless0.001 & E / E \\
NHANES 2017--18 & QSVM - SVM (RBF) & -0.031 {[}-0.083, +0.021{]} & 0.923
& 1.000 & 0.229 & I / I \\
NHANES 2017--18 & QSVM - best (Logistic regression) & -0.076 {[}-0.121,
-0.031{]} & 0.011 & 1.000 & 0.880 & W / W \\
NHANES 2017--18 & VQC (tuned) - best (Logistic regression) & +0.001
{[}-0.018, +0.019{]} & 1.000 & 0.089 & \textless0.001 & I / E \\
\end{longtable}

No quantum model was significantly better than a classical comparator,
and the quantum point estimate exceeded the classical one in only 1 of
12 secondary comparisons. Against the kernel-matched RBF SVM, the QSVM
was equivalent within $\pm$0.05 on 4 datasets (Breast Cancer Wisconsin,
Heart Disease Cleveland, Pima Indians Diabetes and TCGA-BRCA) and within
$\pm$0.02 on 2 (Breast Cancer Wisconsin and TCGA-BRCA), and was not
significantly worse on any dataset. Against the best classical model,
the QSVM was equivalent within $\pm$0.05 on 4 and within $\pm$0.02 on 2 datasets
and significantly worse on NHANES 2017--18; the VQC was equivalent to
the best classical model on 5 datasets at $\pm$0.05 and on 2 at $\pm$0.02.
Equivalence within $\pm$0.01 held in only 1 of the 18 comparisons. Dataset
by dataset, parity therefore holds at the conventional $\pm$0.05 margin but
is rarely established at tighter margins.

Pooled across the six datasets (Figure 2, bottom rows), the QSVM
differed from the kernel-matched RBF SVM by -0.006 AUROC (95\% CI -0.017
to +0.004; τ = 0.005, I² = 32\%), equivalent within $\pm$0.02. Its pooled
difference from the best classical model was -0.013 AUROC (95\% CI
-0.033 to +0.006; τ = 0.011, I² = 67\%), equivalent within $\pm$0.05 but not
$\pm$0.02, and the tuned VQC differed from the best classical model by
-0.007 AUROC (95\% CI -0.019 to +0.004; τ = 0.006, I² = 48\%),
equivalent within $\pm$0.02. No pooled difference significantly favoured a
quantum model, and every pooled point estimate was negative.

Tuning the VQC changed its AUROC by -0.003 to +0.001 relative to the
fixed configuration. The fixed-configuration VQC was equivalent to the
best classical model within $\pm$0.05 on 5 of 6 datasets, against 5 for the
tuned VQC, so the VQC conclusions do not depend on tuning.

The largest deficit against the best classical model, on NHANES 2017--18
(-0.076), shrank to -0.031 and was not significant against the RBF SVM.
Every SVM performed poorly on this dataset relative to logistic
regression, so the deficit appears to reflect the learner more than the
quantum kernel.

\includegraphics[width=0.9\linewidth,keepaspectratio]{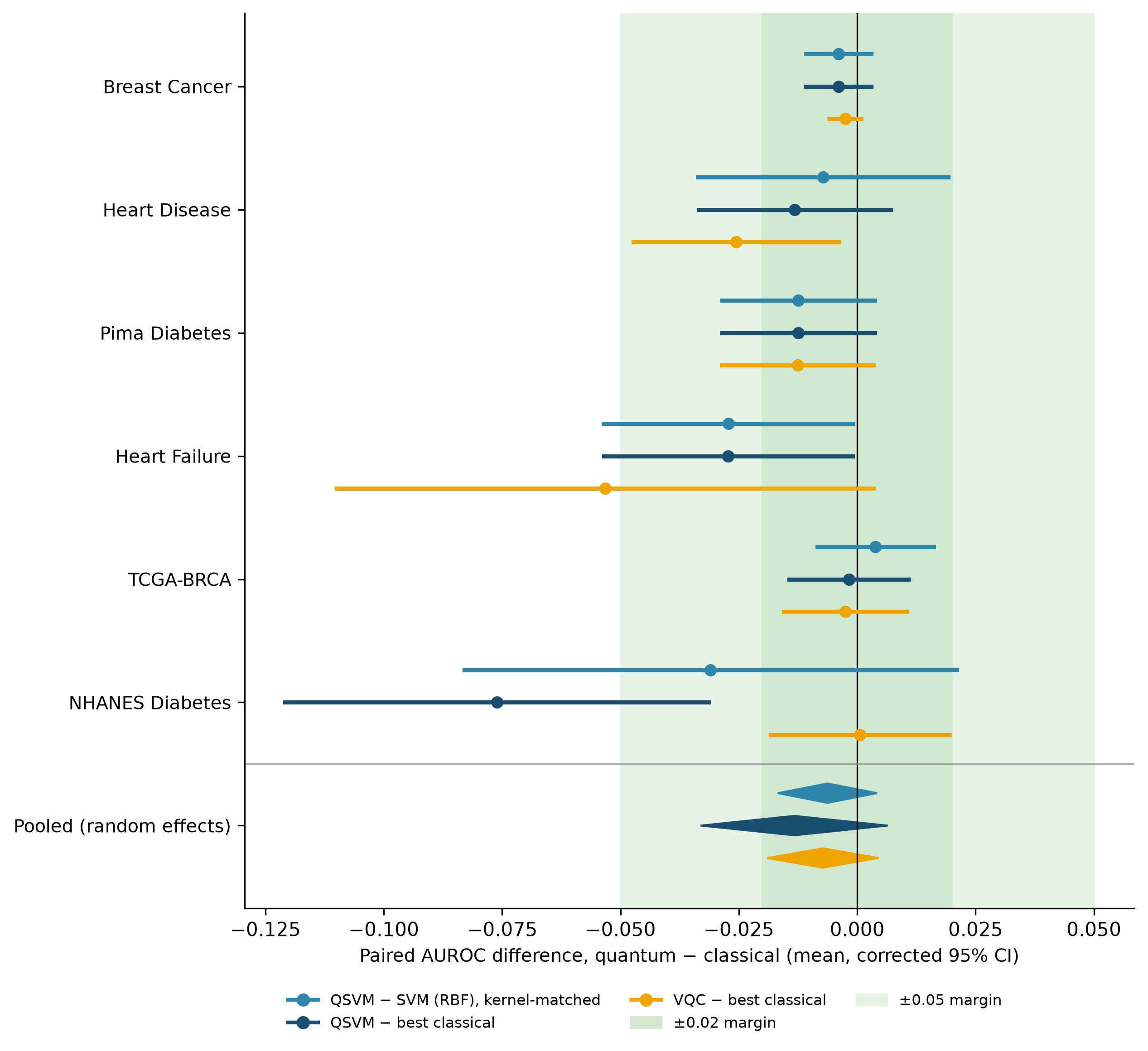}

\emph{Figure 2. Paired AUROC differences with corrected 95\% CIs for
each dataset; diamonds show random-effects pooled estimates with
Hartung--Knapp 95\% CIs. Shaded bands: $\pm$0.02 and $\pm$0.05 equivalence
margins.}

\subsection{5.3 Entanglement and the product-state
kernel}\label{entanglement-and-the-product-state-kernel}

Removing entanglement did not significantly change QSVM AUROC on any
dataset; the product-state kernel had the higher or equal point estimate
on 5 of 6. This kernel is a product of single-qubit overlaps and is
trivially computed classically, yet it came within 0.01 AUROC of the
best classical model on 5 of 6 datasets (Breast Cancer Wisconsin, Heart
Disease Cleveland, Pima Indians Diabetes, Heart Failure and TCGA-BRCA).
The QSVM\textquotesingle s performance on these data therefore did not
depend on entanglement. With all-qubit readout, removing entanglement
from the VQC did not significantly reduce AUROC on any dataset, and the
unentangled VQC had the higher or equal point estimate on 3 of 6.
Details are in Supplementary Table S2.

\subsection{5.4 Noise and finite shots}\label{noise-and-finite-shots}

Depolarising noise at p = 0.01 changed QSVM AUROC by at most 0.007 (not
significant on any dataset), as expected from the near-affine effect of
depolarising noise on the kernel. Larger noise reduced AUROC by up to
0.131 at p = 0.1 (NHANES 2017--18) (Figure 3A). With 1,024 shots per
kernel entry on top of p = 0.01 noise, the QSVM changed by -0.090 to
-0.038 AUROC relative to the noiseless QSVM, and was equivalent to the
best classical model within $\pm$0.05 on none of the six datasets (Figure 3B
shows the shot sweep).

\includegraphics[width=0.9\linewidth,keepaspectratio]{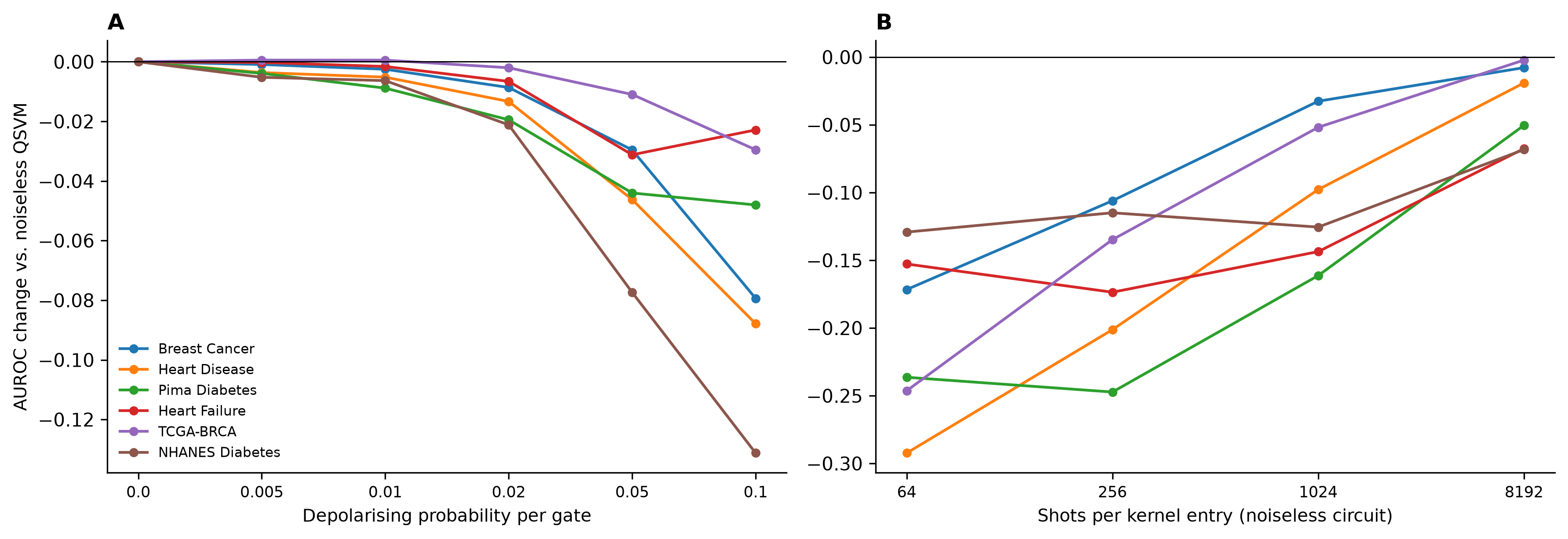}

\emph{Figure 3. Change in QSVM AUROC relative to the noiseless kernel
(first-seed folds, bandwidth and C fixed at the noiseless selection). A:
per-gate depolarising probability. B: shots per kernel entry.}

\subsection{5.5 Kernel regime and data
complexity}\label{kernel-regime-and-data-complexity}

Inner cross-validation chose the grid minimum (0.005) in 7\% of QSVM
fits, a bandwidth of 0.02 or less in 41\%, and never 0.5 or more. At the
modal bandwidth (0.05) the quantum kernel had lower alignment with the
labels than the RBF kernel on all six datasets, with the lowest
alignment on Heart Failure; the geometric difference was 2.42--4.15 at $\lambda=10^{-3}$, small relative to the $\sqrt{n}$ reference scale (17.2--17.3), and the
quantum kernel\textquotesingle s centred alignment with the
best-matching RBF kernel was 0.75--0.82 (Figure 4; Supplementary Table
S3). Alignment and g were computed at the modal bandwidth and a fixed
RBF γ of 1/4, not at each fold's selected settings, so they are
descriptive. The tuned quantum kernels thus operated in the smooth,
small-bandwidth regime in which they approximate classical kernels
{[}22,23{]}, which is consistent with parity and indicates that a
classical kernel can match the quantum one on these data {[}31{]}.

Increasing the number of qubits reproduced exponential concentration of
the untuned kernel {[}20{]}: Breast Cancer Wisconsin: off-diagonal
kernel variance at bandwidth 1 fell from $7.2 \times 10^{-3}$ (4 qubits) to 8.4 ×
10-⁷ (12 qubits), and tuned QSVM AUROC was 0.989 versus 0.994 for the
RBF SVM at 12 qubits; TCGA-BRCA: off-diagonal kernel variance at
bandwidth 1 fell from $7.2 \times 10^{-3}$ (4 qubits) to $7.2 \times 10^{-7}$ (12 qubits),
and tuned QSVM AUROC was 0.954 versus 0.954 for the RBF SVM at 12 qubits
(Supplementary Figure S2).

\includegraphics[width=0.9\linewidth,keepaspectratio]{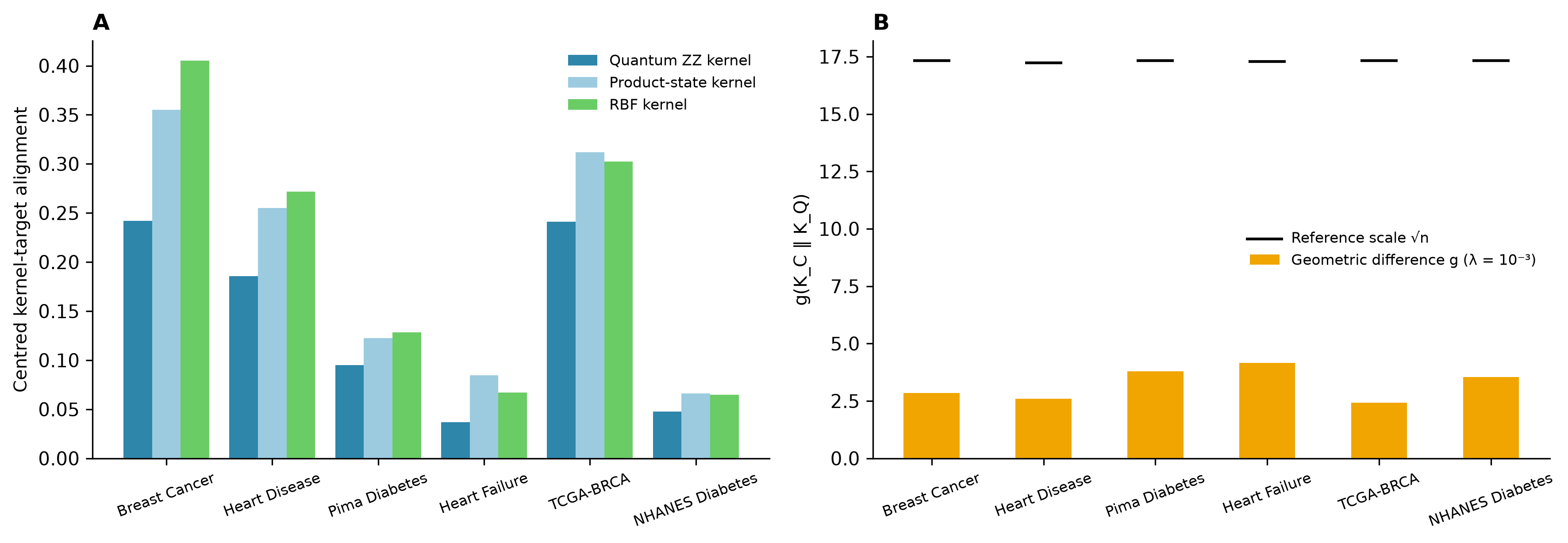}

\emph{Figure 4. A: centred kernel--target alignment. B: geometric
difference g ($\lambda  = 10^{-3}$) with the $\sqrt{n}$ reference scale (not an upper
bound).}

\subsection{5.6 Implementation
sensitivity}\label{implementation-sensitivity}

Fixing the bandwidth at 1 reduced QSVM AUROC from 0.70--0.99 to
0.48--0.59; adding unscaled inputs and the non-entangling CZ--Rz--CZ
block left it at 0.48--0.52 (Figure 5). The tutorial-style configuration
(min--max scaling to {[}0, $\pi${]}, no bandwidth) was less extreme but
still poor, at 0.56--0.89, between 0.02 and 0.24 AUROC below the tuned
QSVM on every dataset. A study using the faulty settings would conclude
that quantum models fail, and one using the tutorial-style settings that
they lag behind classical models on most datasets, whereas the corrected
pipeline gives the results in Table 3; the difference comes from
implementation, not from quantum computing, which is why the Encoding
and Robustness domains of QML-Health-READY are Core items.

\includegraphics[width=0.9\linewidth,keepaspectratio]{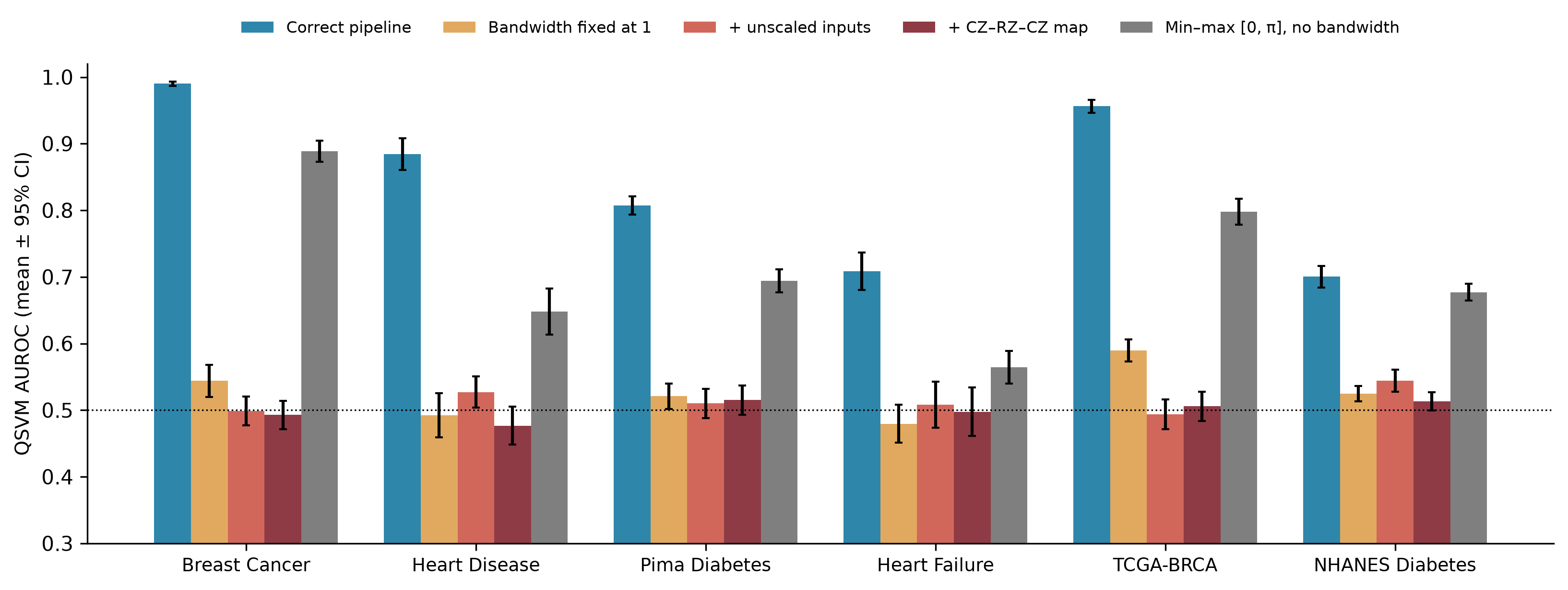}

\emph{Figure 5. QSVM AUROC under cumulative implementation faults (mean
$\pm$ 95\% CI, 25 splits). Dotted line: chance.}

\subsection{5.7 Effect of dimensionality
reduction}\label{effect-of-dimensionality-reduction}

Classical models on all predictors changed AUROC by +0.002 to +0.046
relative to four components (not significant on any dataset after Holm
adjustment). Against the best all-predictor model, the QSVM was
equivalent within $\pm$0.05 on 2 of 6 datasets.

\subsection{5.8 Computational cost}\label{computational-cost}

Per outer split, including tuning, the noiseless QSVM took 0.7--4.2 s
and the noisy QSVM 18--140 s of simulation, versus 0.11--0.25 s for
logistic regression. Simulation time understates hardware cost:
bandwidth tuning on one outer split requires 339,673--15,366,400
distinct kernel-entry circuits (\textbar grid\textbar{} × {[}n(n+1)/2
training + n\_test·n test entries{]}), each executed for hundreds to
thousands of shots, whereas every classical model fits in seconds.

\subsection{5.9 Empirical support for the framework
domains}\label{empirical-support-for-the-framework-domains}

Each QML-Health-READY domain corresponds to an effect measured in this
study, and several were large enough to change a conclusion (Table 5).
The heart-failure comparison is a clean test of the leakage-audit item:
version 2 of our pipeline (Supplementary Table S8) used the same splits
and models but kept follow-up time as a predictor, and every classical
model scored 0.09--0.12 AUROC higher than in the reported analysis.
These observations support the framework's items; they do not replace
the consensus and inter-rater validation that a reporting guideline
requires.

\emph{Table 5. Effects measured in this study that each QML-Health-READY
domain would detect.}

{\def\LTcaptype{none} 
\begin{longtable}[]{@{}
  >{\raggedright\arraybackslash}p{(\linewidth - 4\tabcolsep) * \real{0.1750}}
  >{\raggedright\arraybackslash}p{(\linewidth - 4\tabcolsep) * \real{0.2417}}
  >{\raggedright\arraybackslash}p{(\linewidth - 4\tabcolsep) * \real{0.5833}}@{}}
\toprule\noalign{}
\begin{minipage}[b]{\linewidth}\raggedright
\textbf{Domain}
\end{minipage} & \begin{minipage}[b]{\linewidth}\raggedright
\textbf{Item tested}
\end{minipage} & \begin{minipage}[b]{\linewidth}\raggedright
\textbf{Measured effect in this study}
\end{minipage} \\
\midrule\noalign{}
\endhead
\bottomrule\noalign{}
\endlastfoot
R† Rationale & Data-complexity analysis & Geometric difference 2.4--4.2
against a $\sqrt{n}$ scale of ≈17, and lower quantum than RBF kernel--target
alignment on all six datasets: no data-driven advantage was expected,
and none was seen. \\
E† Encoding & Bandwidth tuning and input scaling & Fixing the bandwidth
at 1 lowered QSVM AUROC by 0.18--0.45, to near chance. \\
A† Algorithm \& baselines & Kernel-matched comparator; entanglement
ablation & The NHANES deficit was -0.076 against the best classical
model but -0.031 (not significant) against the RBF SVM; the
product-state kernel matched the entangling one. \\
D† Data \& splits & Leakage audit (post-baseline predictors) & Including
heart-failure follow-up time inflated classical AUROC by 0.09--0.12. \\
Y† Yield & Pre-specified equivalence margin & QSVM--RBF SVM equivalence
held on 4 of 6 datasets at $\pm$0.05 but 2 at $\pm$0.02 and 0 at $\pm$0.01. \\
R† Robustness & Explicit noise and finite shots & Exact depolarising
noise (p = 0.01) changed AUROC by at most 0.007; 1,024 shots lowered it
by a pooled 0.054. \\
E Evaluation in context & Threshold-dependent metrics & On NHANES the
QSVM reached AUROC 0.70 but MCC 0.002 and balanced accuracy 0.50: at its
default threshold it predicted almost only the majority class. \\
Y Yardstick for cost & Circuit executions per fit & 0.34--15.4 million
kernel-entry circuits per tuned outer split, each repeated for many
shots. \\
\end{longtable}
}

\section{6. Discussion}\label{discussion}
The study explored if quantum classifiers predicted clinical outcomes any better given the identical inputs, splits and tuning budgets as conventional models. They didn’t. In six datasets, well-constructed quantum models were competitive with tuned classical models to within a standard margin but were never better and lost that competitive edge when kernel entries were calculated from a reasonable number of measurements.

\textbf{Principal finding:}
In a matched, paired design, correctly configured quantum classifiers
approached tuned classical models but did not exceed them, and parity
was supported only at the conventional $\pm$0.05 margin. Three observations
are consistent with this. First, inner cross-validation drove the
quantum kernel to small bandwidths, where it behaves like an RBF or
low-order polynomial kernel {[}22,23{]}. Second, the classically trivial
product-state kernel performed as well as the entangling kernel. Third,
the geometric difference was small, so Huang et al.'s necessary
condition for a data-driven advantage was not met {[}31{]}.\\
\textbf{Relation to previous work:} Our results agree with recent independent benchmarks: Kakavand et al.
found no significant advantage in 29 comparisons and quantum-kernel
spectra that were too flat or too concentrated, with the one competitive
configuration costing about 2,000 times more computation {[}15{]};
Leither et al. found no advantage on oncological data {[}16{]}; and
Schnabel and Roth showed, across more than 20,000 quantum-kernel models,
how strongly hyperparameters drive performance {[}33{]}. Several of our
mechanistic findings were anticipated by theory or by benchmarks outside
medicine: the classical character of bandwidth-tuned kernels {[}23{]},
exponential concentration {[}20{]} and the shot cost of kernel
estimation {[}24{]}. What this study adds is their confirmation on
clinical data under a matched, equivalence-tested design; a
kernel-matched comparison showing that apparent quantum deficits can
belong to the learner rather than the kernel; a measured shot penalty
and its interaction with bandwidth tuning; and an experiment showing how
far configuration faults alone move the conclusion.\\
\textbf{Measurement cost:} Finite-shot estimation, not depolarising noise, was the binding
constraint. Exact noise at p = 0.01 changed AUROC by at most 0.007,
whereas adding 1,024 shots per kernel entry lowered it by 0.038--0.090.
The penalty in the shot sweep was largest on the datasets whose tuned
kernels were most compressed (Heart Failure, NHANES 2017--18 and Pima
Indians Diabetes; off-diagonal standard deviation 0.005--0.070), where,
at the bandwidth selected without shot noise, even 8,192 shots left
AUROC 0.05--0.07 below the noiseless value. When bandwidth was re-tuned
under shots, as in the main finite-shot arm, inner cross-validation
chose a larger bandwidth than in the noiseless case in 75\% of fits,
recovering part of the loss. This is consistent with a trade-off between
bandwidth and measurement cost: the small bandwidths that help quantum
kernels generalise leave differences between kernel entries that finite
sampling must resolve {[}20,22,24{]}. Exact state-vector simulation
hides this cost, so finite-shot evaluation belongs in the Robustness
domain of any QML benchmark.\\
\textbf{Comparator choice and implementation sensitivity:} The comparison with the kernel-matched RBF SVM matters. Against the best
of five classical models, a quantum kernel can appear inferior because
SVMs as a class are unsuited to a dataset, as on NHANES 2017--18, where
logistic regression and both tree ensembles beat every SVM. We recommend
that QSVM studies report both comparisons. The
implementation-sensitivity experiment shows that an untuned bandwidth or
unscaled inputs alone produce near-chance results, so published reports
of quantum models failing should be read with the encoding and bandwidth
in mind. Because configuration choices also determine which classical
comparators look strong, the same caution applies to reported successes.\\
\textbf{Scope and clinical implications:}
These conclusions apply to 4-qubit simulations of small, public tabular
datasets (with a 12-qubit check on two datasets). They say nothing about
barren plateaus, which such small circuits cannot exhibit {[}35,37{]},
or about fault-tolerant devices. Where quantum methods may help is
summarised in Supplementary Table S7 and Figure S1: molecular simulation
is the most plausible long-term application, high-dimensional omics a
possible one requiring per-dataset complexity analysis, and routine
tabular risk prediction and imaging unlikely beneficiaries in the NISQ
era. Clinical use of any QML model would also be subject to the
principles of trustworthy healthcare AI {[}27{]} and to medical-device
regulation, and quantum infrastructure is concentrated in a few
high-income settings.\\
\textbf{Recommendations and future:} Authors should give quantum and classical models identical inputs,
splits and tuning budgets; report a kernel-matched classical comparator;
report and check bandwidth selection; verify that noise channels are
active and model finite shots; use split-corrected paired tests with
equivalence margins; and complete the QML-Health-READY checklist.
Reviewers should request it, and funders should prioritise multi-site
and molecular-simulation studies over further single-dataset
classification.\\
Four extensions would test how far these findings generalise. The first
is evaluation on quantum hardware with calibrated noise and error
mitigation, where shot budgets are real rather than simulated. The
second is a larger collection of clinical datasets, including electronic
health record cohorts with external validation, which would allow
cross-dataset inference rather than per-dataset tests. The third is
higher-dimensional omics data with more qubits, the setting our triage
rates as most plausible for a data-driven advantage. The fourth is
validation of QML-Health-READY itself, through a Delphi consensus
process and independent dual-rater application to published studies.

\section{7. Limitations}\label{limitations}
All quantum results come from classical simulation of 4-qubit circuits
(12 qubits in a secondary experiment). Noise was generic depolarising
noise with a binomial shot model that is exact only for the noiseless
estimator, and the noise, shot and qubit-scaling sweeps used the first
seed's folds only; hardware-calibrated noise, readout error and real
hardware were not studied. The analysis was not pre-registered, and
several analyses were added after an initial analysis of an earlier
pipeline. Equivalence margins are conventional rather than clinically
derived, the best classical model was selected post hoc, and
single-model confidence intervals treat dependent splits as independent.
Restricting every model to four principal components ensures a matched
comparison but is clinically artificial, although all-predictor
classical results are reported for context.\\
The datasets are small
public sets without external validation, calibration or decision-curve
analysis, and NHANES survey weights were not applied. The VQC used one
initialisation per split and a small tuning grid (four configurations),
and its entanglement ablation used the fixed configuration. The
bandwidth grid was searched down to 0.005; 7\% of fits chose the
minimum, so for those fits the optimum may lie below the grid. Finally,
the literature synthesis relies on published reviews and benchmarks
rather than a new systematic search, and QML-Health-READY has not been
validated for inter-rater reliability, so a Delphi process and pilot
application to published studies are needed.\\
These limitations mean the
findings should be read as evidence about small, simulated, tabular
clinical settings, not as a verdict on quantum machine learning in
medicine.
The above limitations indicate that the results should be interpreted as insights pertaining to minor, simulated, tabular clinical environments, rather than as a conclusive judgment on the application of quantum machine learning in healthcare.

\section{8. Conclusion}\label{conclusion}
On six clinical datasets, correctly configured quantum classifiers
matched tuned classical models within a conventional margin but never
outperformed them, and their performance did not depend on entanglement.
Their main practical cost was measurement: estimating kernel entries
from a realistic number of shots erased the parity seen in exact
simulation. Implementation choices, not quantum effects, determined
whether they appeared to fail or to succeed. QML-Health-READY sets out
the reporting needed to tell these apart.\\
Future QML studies in health should report matched inputs, a kernel-matched classical comparator, finite-shot estimation and equivalence-tested results, and validation on quantum hardware and larger clinical cohorts is the necessary next step.
\section{Declarations}\label{declarations}
\textbf{Declaration of competing interest:}
The authors declare that they have no known competing financial
interests or personal relationships that could have appeared to
influence the work reported in this paper.\\
\textbf{Data availability}:
All datasets used in this study are publicly available: 
UCI Machine Learning Repository 
(\href{https://archive.ics.uci.edu/}{\textcolor{blue}{archive.ics.uci.edu}}),
UCSC Xena 
(\href{https://xenabrowser.net/}{\textcolor{blue}{xenabrowser.net}}),
and CDC NHANES
(\href{https://wwwn.cdc.gov/Nchs/Data/Nhanes/Public/2017/DataFiles/}{\textcolor{blue}{wwwn.cdc.gov/Nchs/Data/Nhanes/}}).\\
\textbf{Code availability:} The source code and per-split experimental results supporting this study are publicly available on GitHub at
\href{https://github.com/sirajbb/qml-health-ready}{\textcolor{blue}{https://github.com/sirajbb/qml-health-ready}}. An archived version (v3.0.0) is available through Zenodo at
\href{https://doi.org/10.5281/zenodo.23572398}{\textcolor{blue}{https://doi.org/10.5281/zenodo.23572398}}.\\
\textbf{Disclosure of Using AI Tools:}
During the preparation of this work the authors used Claude (Anthropic)
to assist with code development, code review and drafting of the
manuscript. After using this tool, the authors reviewed and edited the
content as needed and take full responsibility for the content of the
published article.\\
\textbf{Ethics statement:} Only public, de-identified data were used; ethics approval was not
required.\\
\textbf{Acknowledgements}: N/A.\\
\textbf{Funding:} N/A.

\section{References}\label{references}

1. Topol EJ. High-performance medicine: the convergence of human and
artificial intelligence. Nat Med. 2019;25(1):44-56.
doi:10.1038/s41591-018-0300-7.

2. Obermeyer Z, Emanuel EJ. Predicting the future --- big data, machine
learning, and clinical medicine. N Engl J Med. 2016;375(13):1216-1219.
doi:10.1056/NEJMp1606181.

3. Preskill J. Quantum computing in the NISQ era and beyond. Quantum.
2018;2:79. doi:10.22331/q-2018-08-06-79.

4. Havlíček V, Córcoles AD, Temme K, Harrow AW, Kandala A, Chow JM,
Gambetta JM. Supervised learning with quantum-enhanced feature spaces.
Nature. 2019;567(7747):209-212. doi:10.1038/s41586-019-0980-2.

5. Schuld M, Killoran N. Quantum machine learning in feature Hilbert
spaces. Phys Rev Lett. 2019;122(4):040504.
doi:10.1103/PhysRevLett.122.040504.

6. Schuld M, Petruccione F. Machine Learning with Quantum Computers. 2nd
ed. Cham: Springer; 2021. doi:10.1007/978-3-030-83098-4.

7. Rebentrost P, Mohseni M, Lloyd S. Quantum support vector machine for
big data classification. Phys Rev Lett. 2014;113(13):130503.
doi:10.1103/PhysRevLett.113.130503.

8. Biamonte J, Wittek P, Pancotti N, Rebentrost P, Wiebe N, Lloyd S.
Quantum machine learning. Nature. 2017;549(7671):195-202.
doi:10.1038/nature23474.

9. Ciliberto C, Herbster M, Ialongo AD, Pontil M, Rocchetto A, Severini
S, Wossnig L. Quantum machine learning: a classical perspective. Proc R
Soc A. 2018;474(2209):20170551. doi:10.1098/rspa.2017.0551.

10. Gupta RS, Wood CE, Engstrom T, Pole JD, Shrapnel S. A systematic
review of quantum machine learning for digital health. npj Digit Med.
2025;8:237. doi:10.1038/s41746-025-01597-z.

11. Bowles J, Ahmed S, Schuld M. Better than classical? The subtle art
of benchmarking quantum machine learning models. arXiv preprint
arXiv:2403.07059. 2024. doi:10.48550/arXiv.2403.07059.

12. Dunjko V, Briegel HJ. Machine learning \& artificial intelligence in
the quantum domain: a review and a roadmap. Rep Prog Phys.
2018;81(7):074001. doi:10.1088/1361-6633/aab406.

13. Tang E. A quantum-inspired classical algorithm for recommendation
systems. In: Proc 51st Annual ACM Symposium on Theory of Computing (STOC
2019). New York: ACM; 2019. p. 217-228. doi:10.1145/3313276.3316310.

14. Sahebi M, Barthe A, Suzuki Y, Holmes Z, Grossi M. On dequantization
of supervised quantum machine learning via random Fourier features.
arXiv preprint arXiv:2505.15902. 2025. doi:10.48550/arXiv.2505.15902.

15. Kakavand S, Strohmeyer C, Schlotter M. Benchmarking quantum kernel
support vector machines against classical baselines on tabular data: a
rigorous empirical study with hardware validation. arXiv preprint
arXiv:2604.18837. 2026. doi:10.48550/arXiv.2604.18837.

16. Leither S, Lubinski T, Kubal M, Johri S. Benchmarking quantum and
classical machine learning models on oncological data. arXiv preprint
arXiv:2608.11373. 2026. doi:10.48550/arXiv.2608.11373.

17. Liu H, Zhang Z, Zheng K, Liang Z, Chen CS, Carothers C, Hahn J.
Quantum machine learning for biomedical classification problems: a
feasibility study on real quantum hardware. Ann Biomed Eng. 2026.
doi:10.1007/s10439-026-04204-8.

18. Priyadharshini M, Raju BD, Banu AF, Kumar PJ, Murugesh V, Rybin O. A
quantum machine learning framework for predicting drug sensitivity in
multiple myeloma using proteomic data. Sci Rep. 2025;15(1):26553.
doi:10.1038/s41598-025-06544-2.

19. Setiawan AE, Rustad S, Syukur A, Soeleman MA, Shidik GF, Mahargya
IL, et al. A systematic literature review of quantum machine learning
for medical: trends, datasets, topics, and methods. Int J Cogn Comput
Eng. 2026;7:609-630. doi:10.1016/j.ijcce.2026.05.002.

20. Thanasilp S, Wang S, Cerezo M, Holmes Z. Exponential concentration
in quantum kernel methods. Nat Commun. 2024;15:5200.
doi:10.1038/s41467-024-49287-w.

21. Shaydulin R, Wild SM. Importance of kernel bandwidth in quantum
machine learning. Phys Rev A. 2022;106(4):042407.
doi:10.1103/PhysRevA.106.042407.

22. Canatar A, Peters E, Pehlevan C, Wild SM, Shaydulin R. Bandwidth
enables generalization in quantum kernel models. Trans Mach Learn Res.
2023. arXiv:2206.06686.

23. Flórez-Ablan R, Roth M, Schnabel J. On the similarity of
bandwidth-tuned quantum kernels and classical kernels. Quantum Sci
Technol. 2025;10(3):035051. doi:10.1088/2058-9565/ade7ad.

24. Miroszewski A, Fellous Asiani M, Mielczarek J, Le Saux B, Nalepa J.
In search of quantum advantage: estimating the number of shots in
quantum kernel methods. arXiv preprint arXiv:2407.15776. 2024.
doi:10.48550/arXiv.2407.15776.

25. Collins GS, Moons KGM, Dhiman P, Wynants L, Myint PK, Majeed A, et
al. TRIPOD+AI statement: updated guidance for reporting clinical
prediction models that use regression or machine learning methods. BMJ.
2024;385:e078378. doi:10.1136/bmj-2023-078378.

26. Mongan J, Moy L, Kahn CE Jr. Checklist for Artificial Intelligence
in Medical Imaging (CLAIM): a guide for authors and reviewers. Radiol
Artif Intell. 2020;2(2):e200029. doi:10.1148/ryai.2020200029.

27. Lekadir K, Frangi AF, Porras AR, Glocker B, Cintas C, Langlotz CP,
et al. FUTURE-AI: international consensus guideline for trustworthy and
deployable artificial intelligence in healthcare. BMJ. 2025;388:e081554.
doi:10.1136/bmj-2024-081554.

28. Schuld M. Supervised quantum machine learning models are kernel
methods. arXiv preprint arXiv:2101.11020. 2021.

29. Abbas A, Sutter D, Zoufal C, Lucchi A, Figalli A, Woerner S. The
power of quantum neural networks. Nat Comput Sci. 2021;1(6):403-409.
doi:10.1038/s43588-021-00084-1.

30. Schuld M, Sweke R, Meyer JJ. Effect of data encoding on the
expressive power of variational quantum-machine-learning models. Phys
Rev A. 2021;103(3):032430. doi:10.1103/PhysRevA.103.032430.

31. Huang HY, Broughton M, Mohseni M, Babbush R, Boixo S, Neven H,
McClean JR. Power of data in quantum machine learning. Nat Commun.
2021;12(1):2631. doi:10.1038/s41467-021-22539-9.

32. Thabet S, Monbroussou L, Mamon EZ, Landman J. When quantum and
classical models disagree: learning beyond minimum norm least square.
npj Quantum Inf. 2026;12(1):81. doi:10.1038/s41534-026-01217-y.

33. Schnabel J, Roth M. Quantum kernel methods under scrutiny: a
benchmarking study. Quantum Mach Intell. 2025;7(1):58.
doi:10.1007/s42484-025-00273-5.

34. Cerezo M, Arrasmith A, Babbush R, Benjamin SC, Endo S, Fujii K, et
al. Variational quantum algorithms. Nat Rev Phys. 2021;3(9):625-644.
doi:10.1038/s42254-021-00348-9.

35. McClean JR, Boixo S, Smelyanskiy VN, Babbush R, Neven H. Barren
plateaus in quantum neural network training landscapes. Nat Commun.
2018;9(1):4812. doi:10.1038/s41467-018-07090-4.

36. Wang S, Fontana E, Cerezo M, Sharma K, Sone A, Cincio L, Coles PJ.
Noise-induced barren plateaus in variational quantum algorithms. Nat
Commun. 2021;12(1):6961. doi:10.1038/s41467-021-27045-6.

37. Cerezo M, Larocca M, García-Martín D, Diaz NL, Braccia P, Fontana E,
et al. Does provable absence of barren plateaus imply classical
simulability? Nat Commun. 2025;16(1):7907.
doi:10.1038/s41467-025-63099-6.

38. Liu Y, Arunachalam S, Temme K. A rigorous and robust quantum
speed-up in supervised machine learning. Nat Phys. 2021;17(9):1013-1017.
doi:10.1038/s41567-021-01287-z.

39. Kübler J, Buchholz S, Schölkopf B. The inductive bias of quantum
kernels. Adv Neural Inf Process Syst. 2021;34:12661-12673.

40. Cortes C, Mohri M, Rostamizadeh A. Algorithms for learning kernels
based on centered alignment. J Mach Learn Res. 2012;13:795-828.

41. Nadeau C, Bengio Y. Inference for the generalization error. Mach
Learn. 2003;52(3):239-281. doi:10.1023/A:1024068626366.

42. Bouckaert RR, Frank E. Evaluating the replicability of significance
tests for comparing learning algorithms. In: Advances in Knowledge
Discovery and Data Mining (PAKDD 2004). Lecture Notes in Computer
Science, vol 3056. Berlin: Springer; 2004. p. 3-12.
doi:10.1007/978-3-540-24775-3\_3.

43. Schuirmann DJ. A comparison of the two one-sided tests procedure and
the power approach for assessing the equivalence of average
bioavailability. J Pharmacokinet Biopharm. 1987;15(6):657-680.
doi:10.1007/BF01068419.

44. Lakens D. Equivalence tests: a practical primer for t tests,
correlations, and meta-analyses. Soc Psychol Personal Sci.
2017;8(4):355-362. doi:10.1177/1948550617697177.

45. Chicco D, Jurman G. Machine learning can predict survival of
patients with heart failure from serum creatinine and ejection fraction
alone. BMC Med Inform Decis Mak. 2020;20(1):16.
doi:10.1186/s12911-020-1023-5.

46. Cancer Genome Atlas Network. Comprehensive molecular portraits of
human breast tumours. Nature. 2012;490(7418):61-70.
doi:10.1038/nature11412.

47. Goldman MJ, Craft B, Hastie M, Repecká K, McDade F, Kamath A, et al.
Visualizing and interpreting cancer genomics data via the Xena platform.
Nat Biotechnol. 2020;38(6):675-678. doi:10.1038/s41587-020-0546-8.

48. National Center for Health Statistics. National Health and Nutrition
Examination Survey (NHANES) 2017-2018 Data Files. Hyattsville, MD: US
Department of Health and Human Services, Centers for Disease Control and
Prevention; 2019. www.cdc.gov/nchs/nhanes.

49. Pedregosa F, Varoquaux G, Gramfort A, Michel V, Thirion B, Grisel O,
et al. Scikit-learn: machine learning in Python. J Mach Learn Res.
2011;12:2825-2830.

50. Bergholm V, Izaac J, Schuld M, Gogolin C, Ahmed S, Ajith V, et al.
PennyLane: automatic differentiation of hybrid quantum-classical
computations. arXiv preprint arXiv:1811.04968. 2018.

51. Benedetti M, Lloyd E, Sack S, Fiorentini M. Parameterized quantum
circuits as machine learning models. Quantum Sci Technol.
2019;4(4):043001. doi:10.1088/2058-9565/ab4eb5.

52. Farhi E, Neven H. Classification with quantum neural networks on
near term processors. arXiv preprint arXiv:1802.06002. 2018.

53. Holm S. A simple sequentially rejective multiple test procedure.
Scand J Stat. 1979;6(2):65-70.

54. DerSimonian R, Laird N. Meta-analysis in clinical trials. Control
Clin Trials. 1986;7(3):177-188. doi:10.1016/0197-2456(86)90046-2.

55. IntHout J, Ioannidis JPA, Borm GF. The Hartung-Knapp-Sidik-Jonkman
method for random effects meta-analysis is straightforward and
considerably outperforms the standard DerSimonian-Laird method. BMC Med
Res Methodol. 2014;14:25. doi:10.1186/1471-2288-14-25.

\end{document}